\documentclass[11pt]{article}

\usepackage[preprint]{acl}

\usepackage{times}
\usepackage{latexsym}

\usepackage[T1]{fontenc}
\usepackage[utf8]{inputenc}

\usepackage{microtype}

\usepackage{inconsolata}

\usepackage{graphicx}
\usepackage{amssymb}
\usepackage{amsmath}
\usepackage{booktabs,multirow,longtable,makecell}
\usepackage{setspace}
\usepackage[table]{xcolor}
\definecolor{lightblue}{RGB}{220,235,255}
\usepackage{subcaption}
\usepackage{enumitem}

\usepackage[most]{tcolorbox}
\newtcolorbox{prompt}[2][]{
    enhanced,
    left=2mm,
    right=2mm,
    top=2mm,
    bottom=2mm,
    boxsep=0.5mm,
    rounded corners,
    title={#2},
    fontupper=\small\linespread{1.25},
    #1
}

\title{CE-RM: A Pointwise Generative Reward Model Optimized via \\ Two-Stage Rollout and Unified Criteria}

\author{
Xinyu Hu$^{1,3}$, Yancheng He$^{3}$, Weixun Wang$^{3}$, Tao Feng$^{3}$, Li Lin$^{1}$, \\ {\bf Jiashun Liu}$^{2,3}$, {\bf Wenbo Su}$^{3}$, {\bf Bo Zheng}$^{3}$, {\bf Xiaojun Wan}$^{1}$\\
$^{1}$Wangxuan Institute of Computer Technology, Peking University \\
$^{2}$Hong Kong University of Science and Technology\!\quad\!$^{3}$Alibaba Group \\
\texttt{\{huxinyu,wanxiaojun\}@pku.edu.cn, \{heyancheng.hyc,weixun.wwx\}@taobao.com}}

\begin{document}
\maketitle
\begin{abstract}
Automatic evaluation is crucial yet challenging for open-ended natural language generation, especially when rule-based metrics are infeasible. Compared with traditional methods, the recent LLM-as-a-Judge paradigms enable better and more flexible evaluation, and show promise as generative reward models for reinforcement learning. However, prior work has revealed a notable gap between their seemingly impressive benchmark performance and actual effectiveness in RL practice. We attribute this issue to some limitations in existing studies, including the dominance of pairwise evaluation and inadequate optimization of evaluation criteria. Therefore, we propose \textbf{CE-RM-4B}, a pointwise generative reward model trained with a dedicated two-stage rollout method, and adopting unified query-based criteria. Using only about 5.7K high-quality data curated from the open-source preference dataset, our CE-RM-4B achieves superior performance on diverse reward model benchmarks, especially in Best-of-N scenarios, and delivers more effective improvements in downstream RL practice.
\end{abstract}

\section{Introduction}

Automatic evaluation has long been a widely studied yet challenging area in natural language processing. Although verifiable tasks such as question answering and mathematics can be evaluated using rule-based metrics, they still face practical difficulties, including diverse answer formats and reasoning process validation. Moreover, there exists a large number of open-ended generation tasks with no well-defined correct answers, such as creative writing and general instruction following. Therefore, it is crucial to develop flexible and efficient automatic evaluation methods.

Traditional research on automatic evaluation has generally evolved through two stages: string-matching metrics such as BLEU~\citep{papineni2002bleu} and ROUGE~\citep{lin2004rouge}, and language-model-based metrics such as BARTScore~\citep{yuan2021bartscore} and COMET~\citep{rei2020comet}. Despite the substantial improvements in evaluation performance, these metrics largely rely on high-quality references and exhibit poor robustness, which can be easily affected by semantic perturbations~\citep{he2023blind}. With the recent rapid development of large language models (LLMs), their improving instruction-following capability has driven a new LLM-based automatic evaluation paradigm (LLM-as-a-Judge)~\citep{wang2023chatgpt, gao2025llm}, which formulates evaluation as an end-to-end generation task to imitate human judgment and supports diverse evaluation functionalities.

Compared with traditional evaluation metrics, this new paradigm achieves performance closer to human-level evaluation. More importantly, it can serve as a generative reward model (GRM)~\citep{mahan2024generative} to provide automatic reward signals for reinforcement learning (RL), especially in scenarios that are difficult to verify with rules. Existing studies~\citep{wang2024self, liu2025inference, DBLP:journals/corr/abs-2505-02387, chen2025judgelrm, guo2025reward, whitehouse2025j1} have made substantial efforts to enhance the performance of GRMs, including distilling data from state-of-the-art LLMs, strengthening the reasoning process during evaluation, and employing rejection sampling for self-bootstrapping. In addition, some work~\citep{liu2024calibrating, xu2024large, liu2025inference} has taken notice of a crucial component of evaluation—criteria—and demonstrated that basing evaluation on high-quality criteria can significantly enhance performance.

Despite the increasing performance of these studies on reward model benchmarks, \citet{DBLP:conf/iclr/WenL0LXLHHZ025} have shown that there remains a notable gap between such benchmark performance and actual effectiveness when deployed in RL. We argue that one primary reason may be that most GRMs adopt a \textbf{pairwise} evaluation protocol. In fact, existing public preference training data are almost pairwise: they are relatively easy to construct and can be generated automatically at scale. In contrast, \textbf{pointwise} data require high-quality human annotations and are thus more expensive. Furthermore, many RM benchmarks are also designed as pairwise comparisons. However, this pairwise protocol is difficult to leverage in RL practice. For example, to apply such RMs in RL with GRPO~\citep{DBLP:journals/corr/abs-2402-03300}, one often needs to compare each pair of trajectories within each rollout group and then use algorithms like Elo~\citep{elo1978rating} to compute a pointwise score for each trajectory, leading to quadratic computational cost and low efficiency.

In addition, although some studies have highlighted the importance of evaluation criteria, they typically treat criteria as part of the evaluation process and optimize the entire evaluation holistically, rather than explicitly strengthening the criteria generation. More notably, their criteria are often generated jointly conditioned on both the query and response. While this approach may yield more comprehensive criteria by incorporating the query requirements and the specific response content, it can lead to inconsistent criteria when multiple responses to the same query are evaluated and compared, thereby introducing biases. This setting is, however, closer to real-world scenarios, such as computing rewards within a rollout group. In Section~\ref{preliminary}, we conduct experiments demonstrating that using unified criteria solely based on the query is increasingly beneficial as the number of responses per query grows.

Motivated by these considerations, we propose a reinforcement learning method with a two-stage rollout—unified \textbf{C}riteria then \textbf{E}valuation—to train our pointwise GRM, \textbf{CE-RM-4B}. Specifically, our model conducts evaluation by first generating evaluation criteria solely based on the query, and then, conditioned on the query and the criteria, producing a detailed evaluation analysis for each response. To train the model, we select about 5.7K data from the public pairwise preference dataset Skywork-Reward-Preference-80K-v0.2~\citep{liu2024skywork} through a series of dedicated quality-filtering procedures. During reinforcement learning, we further design specific algorithms to leverage the original pairwise preference labels to estimate fine-grained reward signals for criteria and evaluation trajectories, respectively, without requiring any additional pointwise score annotations.

We conduct experiments on multiple commonly used RM benchmarks, covering diverse and challenging tasks as well as more realistic Best-of-N scenarios. Experimental results show that, compared with existing work, our CE-RM-4B achieves superior performance with only 4B parameters and less training data. Moreover, on benchmarks where each query is associated with more responses, our model indeed exhibits stronger evaluation capability. We also perform extensive ablation analyses to verify the importance of explicitly optimizing criteria generation and the effectiveness of our proposed methods. Furthermore, we apply our CE-RM-4B in practical RL with GRPO, and observe consistently better improvement. In summary, our contributions are as follows:

\begin{enumerate}
    \item We reveal several limitations of current generative reward models, including the dominance of pairwise evaluation and the lack of dedicated optimization for evaluation criteria.  
    \item We propose a two-stage rollout reinforcement learning method that unifies evaluation criteria conditioned on the query and effectively utilizes pairwise preference labels to train our pointwise GRM, CE-RM-4B. 
    \item Our model achieves superior performance on common reward model benchmarks, especially in more realistic Best-of-N scenarios, and yields more effective improvements in RL practice. Our model is available at \href{https://huggingface.co/PKU-ONELab/CE-RM-4B}{huggingface.co/PKU-ONELab/CE-RM-4B}.
\end{enumerate}

\section{Preliminary Study}
\label{preliminary}

To investigate the role of evaluation criteria and how their different usages affect evaluation performance, we conducted preliminary experiments under three pointwise evaluation settings. For each query $x$ and its candidate responses $\{y_i\}$, the first setting follows the most common practice: the model performs the analysis and then outputs a score (\textit{Direct Evaluation}). It can be formalized as follows, where $r_{\theta}$ denotes the GRM and $e_i$ denotes the evaluation output, $f_{\text{score}}$ is a rule-based function that extracts the score $s_i$ from the evaluation. 
\begin{align}
e_i \sim r_{\theta}\left(x,y_i\right),\!\quad\!\!s_i = f_{\text{score}}\left(e_i\right)
\label{eq1}
\end{align}
In this setting, we do not explicitly require the model to provide evaluation criteria. Nevertheless, the model is allowed to benefit from its reasoning process, and in many cases, it implicitly constructs certain criteria.

The second setting requires the model to first generate explicit evaluation criteria and then perform corresponding analysis and scoring, conditioned on both the query and the response (\textit{Explicit Criteria}), where $c_i$ denotes the generated criteria.
\begin{align}
c_i, e_i \sim r_{\theta}\left(x,y_i\right),\!\quad\!\!s_i = f_{\text{score}}\left(e_i\right)
\label{eq2}
\end{align}
The third setting asks the model to first generate the evaluation criteria solely based on the query, and then evaluate each response conditioned on the query and this \textbf{unified criteria $c$} (\textit{Unified Criteria}), as formalized below.
\begin{align}
c \sim r_{\theta}\left(x\right),\quad\!\!\!\!\!\! e_i \sim r_{\theta}\left(x, y_i, c\right),\quad\!\!\!\!\!\! s_i = f_{\text{score}}\left(e_i\right)
\label{eq3}
\end{align}

We evaluate them on three widely used RM benchmarks: RewardBench~\citep{DBLP:conf/naacl/LambertPMMLCDKZCSH25}, RewardBench2~\citep{malik2025rewardbench}, and RM-Bench~\citep{liurm}, where each query is associated with 2, 4, and 6 responses, respectively. More descriptions are provided in Section~\ref{benchmarks}. We use Qwen3-Max~\citep{qwen3max} as the GRM, and the prompts for the three settings are presented in Figures~\ref{fig:prompt1}--\ref{fig:prompt3} in Appendix~\ref{prompts}. As shown in Table~\ref{tab:preliminary}, explicitly generating criteria consistently improves evaluation performance. Furthermore, as the number of responses per query increases, the advantage of unified criteria becomes more pronounced.

\begin{table}[t]
\centering
\small
\setlength{\tabcolsep}{2.9pt}
\renewcommand{\arraystretch}{1.18}
\begin{tabular}{lccc}
\toprule
\textbf{Evaluation Setting} & \textbf{RWBench} & \textbf{RWBench2} & \textbf{RM-Bench} \\ 
\midrule
Direct Evaluation & 90.0 & 76.7 & 79.8 \\
Explicit Criteria & 90.3 & 78.8 & 81.5 \\
Unified Criteria & 90.6 & 79.8 & 83.0 \\
\bottomrule
\end{tabular}
\caption{The results of Qwen3-Max with three different evaluation settings on three reward model benchmarks: RewardBench (RWBench), RewardBench2 (RWBench2) and RM-Bench.}
\label{tab:preliminary}
\end{table}

\section{Methodology}

Motivated by the above exploration, we aim to train a pointwise GRM with unified criteria, whose evaluation process follows Eq.~(\ref{eq3}), while explicitly strengthening the criteria generation capability. To this end, we propose a specific reinforcement learning method based on a two-stage rollout. In addition, we design algorithms to convert pairwise preference labels into fine-grained rewards for criteria and evaluation trajectories, respectively.

\subsection{Data Curation}

To obtain training data with high quality and instance efficiency, we start from a widely used preference dataset, Skywork-Reward-Preference-80K-v0.2, and perform data filtering. Each instance contains a query and a pair of chosen and rejected responses $(x, y^c, y^r)$, corresponding to a pairwise preference label. We first use Qwen3-4B-Instruct-2507~\citep{qwen3technicalreport} to perform multiple pointwise evaluations for each instance, and compute accuracy as the fraction of trials in which the chosen response $y^c$ is assigned a higher score than the rejected response $y^r$. We then retain about 29K instances whose accuracy does not exceed 0.6, representing those the model is uncertain about or judges incorrectly. Furthermore, to ensure the diversity of queries, we employ Qwen3-Max to identify the task type of each query and cluster them using Qwen3-4B-embedding~\citep{qwen3embedding}. Finally, we perform stratified random sampling to obtain approximately 12K instances whose task-type distribution is as balanced as possible.

\subsection{Cold Start}
\label{cold_start}
We distill a small amount of supervised fine-tuning (SFT) data from Qwen3-Max for cold-starting our reward model under the two-stage evaluation protocol. For each query $x$, following the prompt in Figure~\ref{fig:prompt3} and Eq.~(\ref{eq3}), we require Qwen3-Max to first generate three sets of criteria $\{c_{i}\}_{i=1}^{3}$ solely based on the query. Then, conditioned on $x$ and each $c_{i}$, the model generates three evaluations for the chosen response $y^{c}$ and the rejected response $y^{r}$, denoted as $\{e^{c}_{ij}\}_{j=1}^3$ and $\{e^{r}_{ij}\}_{j=1}^3$, respectively.

In particular, the evaluation is conducted based on each criterion item in the generated criteria set with the corresponding sub-score, and then the overall analysis and final score are derived based on these contents to ensure reliability.
Moreover, to prevent the criteria set from failing to cover specific aspects of a response, we allow the model to introduce an additional adjustment beyond the criteria when necessary, as shown in stage 2 of Figure~\ref{fig:prompt3}.
The score range is set to 0–10 with a half-point increment to mitigate the limited granularity of pointwise scoring. In preliminary tests, this setting achieves a good balance between granularity and reliability, while finer granularity, such as 0.1, often causes the scores to deviate from well-grounded evidence.

Since each instance only contains the preference label and no human evaluation as a gold standard, we design empirical measures to assess the quality of the instance, criteria, and evaluation, making the cold-start data as reliable as possible.
\paragraph{Instance} We select the instance where, based on all three criteria sets, the corresponding evaluations consistently yield the higher scores for the chosen response than the rejected response, ensuring the correctness of the preference.
\begin{align}
s^c_{ij} = f_{\text{score}}\left(e^{c}_{ij}\right)&,\quad \!\!\! s^r_{ik} = f_{\text{score}}\left(e^{c}_{ik}\right) \\
\prod_{1 \le i \le 3}\prod_{1 \le j, k \le 3}
\mathbb{I} & \left(s^{c}_{ij} > s^{r}_{ik}\right) = 1
\end{align}
\paragraph{Criteria} For each retained instance above, among its three criteria sets, we select $e_{i^*}$ where the variance of its induced evaluation scores is the smallest. If this variance exceeds a threshold, the instance is discarded. This aims to ensure that the criteria are unambiguous, enabling stable and correct scoring during evaluation.
\begin{align}
    i^* = \arg\min_{1 \le i \le 3}\Bigl(&
 \operatorname{Var}\left(\{s^c_{ij}\}_{j=1}^{3}\right) \notag\\
& + \operatorname{Var}\left(\{s^r_{ij}\}_{j=1}^{3}\right)
\Bigr)
\end{align}
\paragraph{Evaluation} For each selected instance and its criteria set, we choose the response evaluation whose score is the median among the three. Furthermore, to reuse these data in subsequent reinforcement learning while preventing the model from directly observing the preference label, we keep only one response (either chosen or rejected) along with its corresponding evaluation. And we design a specific algorithm for the aforementioned response selection to make the resulting score distribution as uniform as possible.

Finally, we obtain about 2.2K instances, denoted as $\mathcal{D}_{\mathrm{SFT}}\!=\!\{(x_n,y_n,c_n,e_n)\}_{n=1}^{N}$, where each contains a query, a response, a criteria set, and an evaluation. The cold-start training objective is defined as follows. For each instance, we simultaneously optimize the generation of criteria and evaluation, but under different conditioning contexts.
\begin{align}
\mathcal{L}_{\mathrm{SFT}} &
= - \frac{1}{N}\sum_{n=1}^{N}\Big(
\sum_{t}
\log r_{\theta}\!\left(c_{n,t}\mid x_n, c_{n,<t}\right) \notag\\
& +\sum_{t}
\log r_{\theta}\!\left(e_{n,t}\mid x_n,y_n,c_n, e_{n,<t}\right)
\Big)
\end{align}

\begin{table*}[t]
\centering
\small
\setlength{\tabcolsep}{2.4pt}
\renewcommand{\arraystretch}{1.196}
\begin{tabular}{lccccccc}
\toprule
\textbf{Reward Model} & \textbf{Data} & \textbf{RWBench} & \textbf{RWBench2} & \textbf{RM-Bench} & \textbf{PPE Corr} & \textbf{JudgeBench} & \textbf{Overall}\\

\midrule
\multicolumn{8}{l}{\emph{Pairwise RM}} \\

GPT-4o~\citep{DBLP:journals/corr/abs-2410-21276} & - & 86.7 & - & 72.5 & 57.6 & 56.6 & - \\
Llama-3.3-70B-Instruct & - & 85.4 & - & 69.5 & 65.7 & 48.6 & - \\
JudgeLRM-7B~\citep{chen2025judgelrm} & 100K & 75.2 & - & 58.7$^*$ & 42.6$^*$ & 54.6$^*$ & - \\
RM-R1-Qwen-7B~\citep{DBLP:journals/corr/abs-2505-02387} & 73K & 85.2 & - & 70.2 & 57.9$^*$ & 59.1$^*$ & - \\
RM-R1-Qwen-32B~\citep{DBLP:journals/corr/abs-2505-02387} & 73K & 91.4 & - & 79.1 & 59.3$^*$ & 68.6$^*$ & - \\
RRM-7B~\citep{guo2025reward} & 420K & 82.2 & - & 70.8$^*$ & 60.1$^*$ & 60.9$^*$ & - \\
RRM-32B~\citep{guo2025reward} & 420K & 91.2 & - & \underline{82.8$^*$} & 67.9$^*$ & \underline{75.1$^*$} & - \\
RISE-Judge-7B~\citep{yu2025improve} & 73K & 88.2 & - & 68.0$^*$ & 60.4$^*$ & 59.7$^*$ & - \\
RISE-Judge-32B~\citep{yu2025improve} & 40K & 92.7 & - & 72.3$^*$ & 55.0$^*$ & 66.3$^*$ & - \\
Think-J-32B~\citep{huang2025think} & 9.8K & 90.5 & - & 79.8 & - & - & - \\
CompassJudger2-7B~\citep{zhang2025compassjudger} & ? & 91.0 & - & 71.5$^*$ & 60.2$^*$ & 63.1 & -\\
CompassJudger2-32B~\citep{zhang2025compassjudger} & ? & 92.6 & - & 73.2$^*$ & 56.6$^*$ & 65.5 & -\\
J1-Llama-8B~\citep{whitehouse2025j1} & 22K & 85.7 & - & 73.4 & 59.2 & 42.0 & - \\
J1-Llama-70B~\citep{whitehouse2025j1} & 22K & \underline{\textbf{93.3}} & - & 82.7 & \underline{72.9} & 60.0 & - \\
\midrule
\multicolumn{8}{l}{\emph{Pointwise RM}} \\
Gemini-2.5-Flash~\citep{DBLP:journals/corr/abs-2507-06261} & - & 80.7 & \underline{75.9} & 70.8 & 61.9 & 66.9 & 71.2 \\
CompassJudger1-32B~\citep{cao2024compassjudger} & 900K & 81.2$^*$ & 56.5$^*$ & 54.4$^*$ & 48.0$^*$ & 42.3$^*$ & 56.5$^*$ \\
DeepSeek-GRM-27B~\citep{liu2025inference} & 237K & 86.0 & - & - & 59.8 & - & - \\
CLoud-Gemma-2-27B~\citep{liu2025inference} & 237K & 82.0 & - & - & 62.4 & - & - \\
J1-Llama-8B~\citep{whitehouse2025j1} & 22K & - & - & - & 53.8 & 57.7 & - \\
J1-Llama-70B~\citep{whitehouse2025j1} & 22K & - & - & - & 65.0 & 60.0 & - \\
TIR-Judge-Distill-4B~\citep{xu2025incentivizing} & 26K & 76.6 & 67.3 & 71.9 & 65.9 & 66.7 & 69.7 \\
TIR-Judge-Zero-4B~\citep{xu2025incentivizing} & 26K &  77.3 & 68.3 & 72.8 & 69.8 & 70.4 & 71.7 \\
TIR-Judge-Distill-8B~\citep{xu2025incentivizing} & 26K & 81.0 & 71.6 & 76.7 & 71.0 & 68.2 & 73.7\\
TIR-Judge-Zero-8B~\citep{xu2025incentivizing} & 26K & 81.4 & 73.4 & 76.3 & 70.3 & 67.5 & 73.8\\

\rowcolor{lightblue} CE-RM-4B (Ours) & 5.7K & 89.0 & 74.6 & 79.8 & 69.7 & 73.7 & \underline{77.4} \\
\rowcolor{lightblue} CE-RM-4B (Scaling@2) & 5.7K & 89.4 & 75.9 & 81.9 & 73.1 & 75.7 & 79.2 \\
\rowcolor{lightblue} CE-RM-4B (Scaling@4) & 5.7K & 90.0 & \textbf{76.3} & \textbf{83.2} & \textbf{75.0} & \textbf{76.3} & \textbf{80.2} \\

\bottomrule
\end{tabular}
\caption{The results of our CE-RM-4B compared with other baselines on five common reward model benchmarks, as well as their overall performance. RWBench, RWBench2, and PPE Corr represent the RewardBench, RewardBench2, and PPE Correctness, respectively. And scaling@k denotes the performance obtained by applying test-time scaling for k times and using the averaged evaluation scores. The results marked with $^*$ are those not reported in the original paper but tested and obtained by ourselves; The results highlighted in \textbf{bold} and \underline{underlining} indicate the best one in each column, and the latter refers specifically to comparisons conducted without test-time scaling.}
\label{tab:main_result}
\end{table*}

\subsection{RL with Two-Stage Rollout}

Unlike the cold start, reinforcement learning requires data in the original triplet form $(x, y^c, y^r)$. Since criteria and evaluations are not needed, we relax the filtering condition used in cold start, and keep instances where at least one set of criteria leads to entirely correct evaluations as follows, resulting in about 5.7K instances, denoted as $\mathcal{D}_{\mathrm{RL}}$.
\begin{align}
\sum_{1 \le i \le 3} \prod_{1 \le j, k \le 3}
\mathbb{I}\!\left( s^{c}_{ij} > s^{r}_{ik}\right) \ge 1
\end{align}

The evaluation process conducted by our reward model consists of two stages: criteria generation and evaluation generation, both of which need to be optimized. Therefore, building upon the reinforcement learning with GRPO~\citep{DBLP:journals/corr/abs-2402-03300}, we adopt a two-stage rollout scheme. Similar to the procedure described in Section~\ref{cold_start}, for each query, the model first generates $n_c$ criteria trajectories in the first stage; then, conditioned on the query and each generated criteria trajectory, it produces $n_e$ evaluation trajectories for the chosen and rejected responses, respectively, in the second stage.

Given that only pairwise preference labels are available, we design the following methods to estimate fine-grained reward signals for criteria and evaluation trajectories, improving both reliability and discriminability. The reward for each set of criteria $c_i$ is defined as the win rate of pairwise comparison between the chosen and rejected responses in terms of their induced scores from the subsequent corresponding evaluations, estimated in an aggregated manner to reduce bias.
\begin{align}
    \mathcal{R}_{\mathrm{criteria}} = \frac{1}{n_e^2}\sum_{1 \le j, k \le n_e}
\mathbb{I} & \left(s^{c}_{ij} > s^{r}_{ik}\right)
\end{align}
The reward for each evaluation $e_{ij}^c$ of the chosen response $y^c$ is defined as its win rate in pairwise score comparisons against all evaluations of the rejected response, multiplied by a binary format reward (0 for incorrect, 1 otherwise); and the reward for each evaluation $e_{ij}^r$ of $y^r$ is defined analogously.
\begin{align}
    \mathcal{R}_{\mathrm{eval}}^c = \frac{1}{n_e}\sum_{1 \le k \le n_e}\!
\mathbb{I} & \left(s^{c}_{ij} > s^{r}_{ik}\right) \cdot \mathcal{R}_{\mathrm{format}}
\end{align}

Notably, in this two-stage rollout, the criteria and the evaluations of chosen and rejected responses are all treated as optimization objectives for the current training instance, but they correspond to three different conditional generations. Therefore, unlike typical GRPO, which groups rollouts by instance, we partition trajectories into three sub-groups (criteria sets, chosen-response evaluations, rejected-response evaluations) for subsequent relative advantage estimation and optimization.

\section{Experiments}

\subsection{Implementation Details}
\label{implementation}

We use Qwen3-4B-Instruct-2507~\citep{qwen3technicalreport} as the foundation model for training, to balance the model capability and experimental cost. For the cold start, we set the batch size to 64, the learning rate to 1e-5, and train for 3 epochs. For reinforcement learning, the mini-batch size is 64, with 4 mini-batches per step; the learning rate is 2e-6, and the KL coefficient is 1e-3.
More concretely, for each query, we generate four sets of criteria in the first-stage rollout ($n_c$ = 4); in the second stage, we generate two evaluations for both the chosen and rejected responses ($n_r$ = 2). Consequently, each instance involves 4 criteria trajectories and 16 evaluation trajectories, i.e., 20 trajectories in total. Moreover, our reinforcement learning is implemented based on the open-source ROLL framework~\citep{DBLP:journals/corr/abs-2506-06122}. During inference, we set the temperature to 0 to improve determinism.

\subsection{Benchmarks}
\label{benchmarks}
To comprehensively evaluate the performance of our model, we adopt a collection of widely used reward model benchmarks, covering diverse task types, challenging settings, and Best-of-N scenarios. Specifically, we select:
\paragraph{RewardBench~\citep{DBLP:conf/naacl/LambertPMMLCDKZCSH25}} One of the earliest and most popular benchmarks for reward models, spanning multiple categories such as chat, safety, and reasoning. However, its difficulty may be relatively low for current models.
\paragraph{RewardBench2~\citep{malik2025rewardbench}} A substantial upgrade in both difficulty and diversity compared with RewardBench. It features more realistic human queries and a global Best-of-N evaluation protocol, and has been shown to correlate more strongly with downstream RLHF performance.
\paragraph{RM-Bench~\citep{liurm}} Focuses on robustness under subtle content differences and stylistic biases, testing whether a reward model truly judges based on content rather than style.
\paragraph{PPE Correctness~\citep{frickevaluate}} Built upon verifiable and challenging benchmarks (MATH, MBPP, MMLU-Pro, GPQA, IFEval) to test alignment in Best-of-N tasks. It has likewise been validated to reflect RLHF performance.
\paragraph{JudgeBench~\citep{tanjudgebench}} Derived from a set of real-world and difficult evaluation datasets (e.g., MMLU-Pro, LiveBench). It assesses the capability to discriminate answer correctness through pairwise comparison.

\begin{table}[t]
\centering
\small
\setlength{\tabcolsep}{5pt}
\renewcommand{\arraystretch}{1.18}
\begin{tabular}{cccc}
\toprule
\textbf{Trajectory} & \textbf{Criteria ($n_c$)} & \textbf{Evaluation ($n_e$)} & \textbf{Overall} \\ 
\midrule
9 & 1 & 4 & 72.7 \\
10 & 2 & 2 & 74.1 \\
12 & 4 & 1 & 73.6\\
18 & 2 & 4 & 74.4 \\
21 & 3 & 3 & 75.3 \\
20 & 4 & 2 & 77.4 \\
\bottomrule
\end{tabular}
\caption{The overall performance of the reward model trained with different combinations of two-stage rollout numbers on five benchmarks. The columns of trajectory, criteria, and evaluation represent the number of total trajectories, generated criteria trajectories and evaluation trajectories for each instance during RL, respectively.}
\label{tab:rollout_numbers}
\end{table}

\subsection{Main Results}

In Table~\ref{tab:main_result}, we compare our CE-RM-4B with a great range of latest generative reward models and several general-purpose LLMs, which are grouped into two categories by evaluation protocol: pairwise and pointwise. Since many prior works do not report results on all benchmarks, we test those open-source reward models ourselves and supplement the results, which are marked with $^*$ in the table. Moreover, as RewardBench2 is not compatible with pairwise reward models by design, corresponding results are unavailable.

The experimental results show that our CE-RM-4B achieves superior overall evaluation performance, despite using much less training data and a smaller model size than other baselines. Its advantage is particularly pronounced on RewardBench2 and PPE Correctness, which are aligned with Best-of-N scenarios, as well as on RM-Bench, where each query is associated with a larger number of responses. It highlights the benefit of our enforcing unified query-based evaluation criteria.

\begin{figure}[t]
    \centering
    \includegraphics[width=0.99\linewidth]{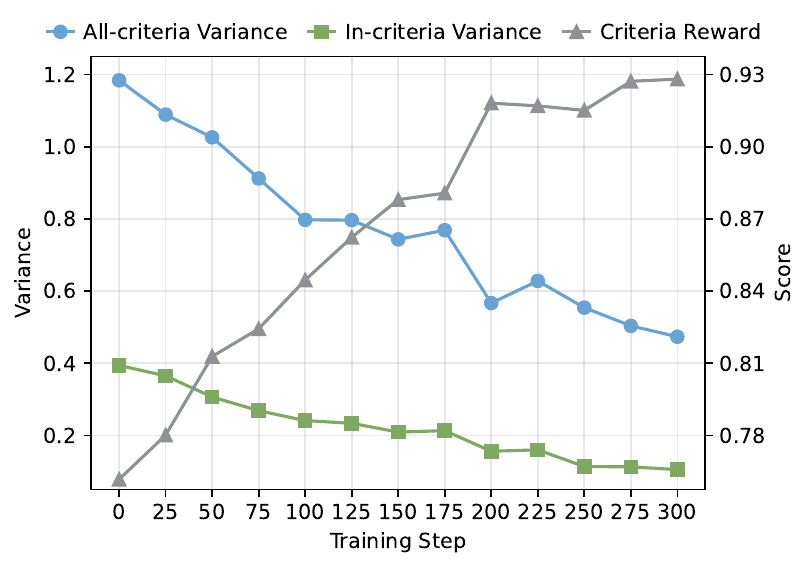}
    \caption{The variance of the evaluation scores corresponding to the same criteria trajectory (in-criteria), and to all criteria trajectories (all-criteria) for each instance, as well as the criteria rewards during RL.}
    \label{fig:variance}
\end{figure}

\begin{figure*}[t]
    \centering
    \includegraphics[width=0.99\linewidth]{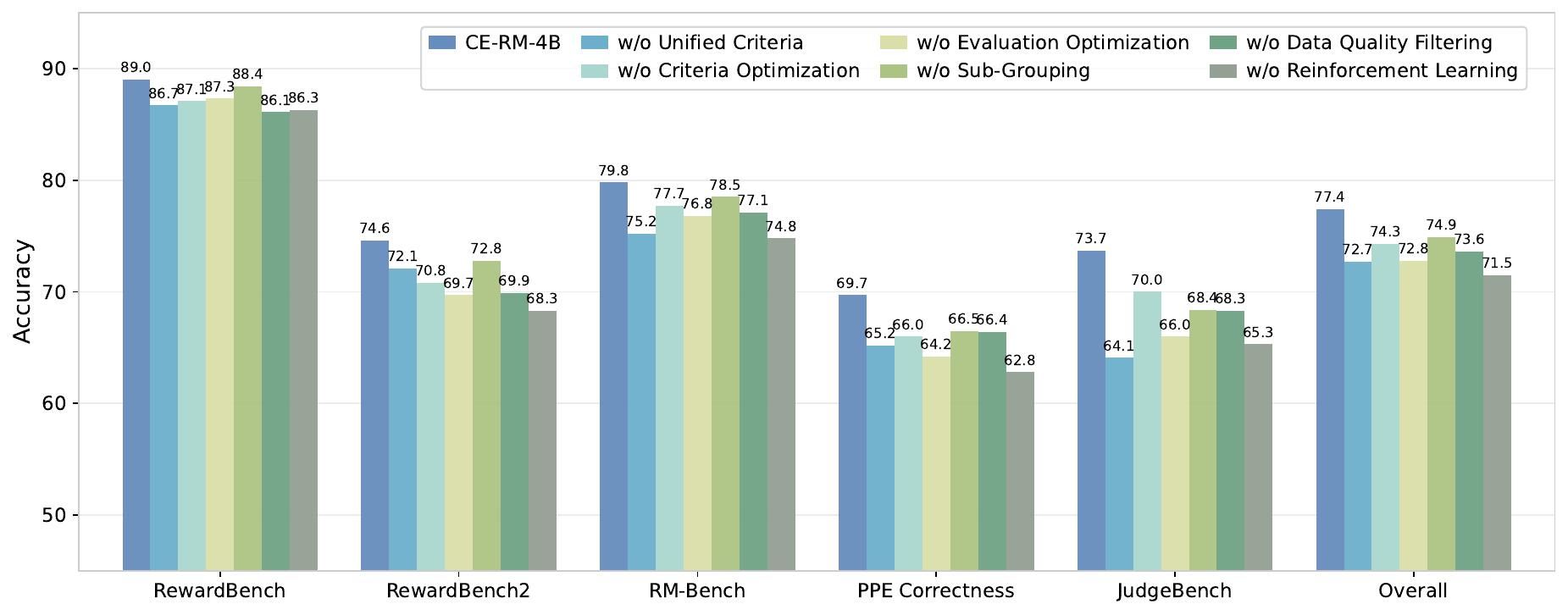}
    \caption{The results of the ablation experiments that compare our CE-RM-4B with other reward models trained without different key components of our methods.}
    \label{fig:ablation}
\end{figure*}

\subsubsection{Test-Time Scaling}
Similar to work such as~\citet{liu2025inference}, our model can be improved via test-time scaling using averaged scores from multiple inferences. Since pointwise scores from GRMs often have limited granularity, ties occur commonly and can degrade evaluation performance. And averaging multiple sampled scores effectively mitigates this issue. Table~\ref{tab:main_result} also reports results of scaling at 2 and 4 times. The gains are most evident on RM-Bench, PPE Correctness, and JudgeBench, where ties are not explicitly handled. Moreover, owing to its relatively fewer parameters, our model incurs low computational overhead for test-time scaling, and it even outperforms other substantially larger reward models that employ pairwise evaluation protocols.

\subsubsection{Two-Stage Rollout Combination}
We further explore combinations of trajectory numbers in our two-stage rollout method. As shown in Table~\ref{tab:rollout_numbers}, increasing total trajectories for each instance generally leads to better performance. Under comparable trajectory budgets, adopting 4 criteria trajectories and 2 evaluation trajectories yields the best result, suggesting that while maintaining a certain number of evaluations to stabilize the estimation of reward scores, optimizing the criteria is more critical, which confirms our motivation.

\subsubsection{Evaluation Robustness}
In Figure~\ref{fig:variance}, we show the variance of scores from the evaluations based on the same criteria trajectory (in-criteria), and on all criteria trajectories (all-criteria) for each instance during RL. Benefiting from the effective optimization of both the criteria and evaluation, our model produces increasingly stable evaluation scores as training proceeds. Similar results are also observed on the test benchmarks, as shown by the comparisons in Table~\ref{tab:robustness}, which improves the robustness of CE-RM-4B in practical deployments, such as the low-cost or low-latency settings where only a single inference is allowed.

\begin{table}[t]
\centering
\small
\setlength{\tabcolsep}{4.4pt}
\renewcommand{\arraystretch}{1.152}
\begin{tabular}{lccc}
\toprule
\textbf{Model} & \textbf{Intra} & \textbf{Inter} & \textbf{Acc} \\
\midrule
Qwen3-4B-Instruct-2507 & 0.621 & 1.423 & 72.8 \\
\quad+ Cold Start & 0.482 & 0.893 & 76.5 \\
CE-RM-4B w/o Optimized Criteria & 0.227 & 0.515 & 80.0 \\
CE-RM-4B (Ours) & 0.181 & 0.441 & 81.1 \\
\bottomrule
\end{tabular}
\caption{The score variance and accuracy of different models averaged across RewardBench, RewardBench2, and RM-Bench. Intra and Inter denote the variance of the evaluation scores across four samplings, where the former is computed from four evaluations generated based on one sampled criteria set, whereas the latter is calculated by sampling four different sets of criteria and the corresponding evaluations respectively. Specifically, CE-RM-4B w/o Optimized Criteria refers to a compositional setting where the criteria are generated by the RM trained only with the cold start, and the evaluations are then produced based on them 
by our CE-RM-4B.}
\label{tab:robustness}
\end{table}

\section{Ablation Analyses}
\label{ablation}
We further conduct ablation studies on key components of our method to validate their importance, whose training settings are similar to Section~\ref{implementation}:
\paragraph{Unified Criteria} This is the most critical component and one of our starting points. For comparison, we train another reward model using the traditional scheme that generates criteria and evaluation jointly conditioned on the query and response (i.e., Eq.~(\ref{eq2})), with a similar data construction and total trajectory number during the RL procedure.
\paragraph{Criteria Optimization} In the two-stage rollout during reinforcement learning, after computing the reward signals, we discard criteria trajectories in the optimization objective to verify the importance of explicitly enhancing criteria generation.
\paragraph{Evaluation Optimization} Similarly, we discard evaluation trajectories in the optimization objective, to examine the influence of optimizing only the criteria.
\paragraph{Sub-Grouping} We replace our sub-grouping strategy with the conventional grouping over all trajectories within the rollout of each instance.
\paragraph{Data Quality Filtering} We significantly simplify the quality filtering pipelines used in both cold start and RL, retaining only basic correctness checks on overall judgments, to demonstrate the necessity of our carefully designed quality estimation.
\paragraph{Reinforcement Learning} We also evaluate the model trained only with the cold start to quantify the improvements brought by RL.

Figure~\ref{fig:ablation} presents the results of the ablation experiments above, highlighting the effectiveness of the unified query-based criteria and our proposed training methodology.

\begin{table*}[t]
\centering
\small
\setlength{\tabcolsep}{8.7pt}
\renewcommand{\arraystretch}{1.2}
\begin{tabular}{lcccccc}
\toprule
\multirow{2}{*}{\raisebox{-3.2pt}{\textbf{Model}}} & \multicolumn{2}{c}{\textbf{Arena-Hard v0.1}} & \multicolumn{2}{c}{\textbf{v2 Hard Prompt}} & \multicolumn{2}{c}{\textbf{v2 Creative Writing}} \\

\cmidrule(lr){2-3} \cmidrule(lr){4-5} \cmidrule(lr){6-7} 
 & \multicolumn{1}{c}{Score} & \multicolumn{1}{c}{CI}
 & \multicolumn{1}{c}{Score} & \multicolumn{1}{c}{CI}
 & \multicolumn{1}{c}{Score} & \multicolumn{1}{c}{CI} \\
 
\midrule
Qwen3-8B & 66.5 & (-2.7 / +2.1) & 9.8 & (-1.0 / +0.8) & 14.4 & (-1.4 / +1.8) \\
Qwen3-14B & 77.4 & (-2.6 / +1.4) & 17.1 & (-1.8 / +1.4) & 24.7 & (-2.1 / +2.0) \\
\midrule
\emph{GRPO with group size = 4} \\
Qwen3-8B + CompassJudger1-32B & 75.0 & (-2.2 / +2.0) & 13.4 & (-0.8 / +1.1) & 27.1 &  (-2.6 / +2.3) \\
Qwen3-8B + RM w/o unified criteria & 71.0 & (-2.5 / +2.0) & 12.9 & (-1.2 / +1.0) & 37.6 & (-2.8 / +2.5) \\
Qwen3-8B + \textbf{CE-RM-4B} (Ours) & 75.7 & (-1.5 / +2.0) & 16.3 & (-1.4 / +1.3) & 42.9 & (-2.2 / +2.3) \\
Qwen3-8B + \textbf{CE-RM-4B} (Scaling@4) & \textbf{78.3} & (-1.6 / +1.4) & 17.6 & (-1.2 / +1.4) & 48.0 & (-2.2 / +2.1) \\
\midrule
\emph{GRPO with group size = 8} \\
Qwen3-8B + CompassJudger1-32B & 74.7 & (-1.8 / +1.8) & 13.6 &  (-1.0 / +1.1) & 39.4 &  (-2.1 / +2.4) \\
Qwen3-8B + RM w/o unified criteria & 72.1 & (-2.4 / +1.7) & 13.5 & (-1.4 / +1.2) & 39.9 & (-2.1 / +2.3) \\
Qwen3-8B + \textbf{CE-RM-4B} (Ours) & 77.6 & (-1.6 / +1.7) & \textbf{18.2} & (-1.6 / +1.4) & \textbf{49.1} & (-2.1 / +1.9) \\
\bottomrule
\end{tabular}
\caption{Evaluation results on Arena-Hard for the original policy models and models trained with different optimizations. The three columns correspond to Arena-Hard v0.1 and the two categories of Arena-Hard v2, evaluating with the official configuration and style control.}
\label{tab:RL_practice}
\end{table*}

\section{RL Practice}

Since most generative reward models adopt the pairwise evaluation, their downstream RLHF performance is typically validated by preference label annotation and DPO-style training~\citep{rafailov2023direct}. To better approximate real-world scenarios, we instead adopt on-policy reinforcement learning, using the evaluation scores from RMs as the reward signal to optimize the policy model with GRPO. In particular, we randomly sample 50K queries from Tulu 3 dataset~\citep{lambert2025tulu3pushingfrontiers} as training data. And we evaluate the effectiveness of policy optimization using Arena Hard~\citep{li2024crowdsourced, arenahard2024}, a popular automatic evaluation arena that contains challenging real-world user queries and employs powerful LLMs as the judge, showing strong correlation with human judgments.

\begin{figure}[t]
    \centering
    \includegraphics[width=0.98\linewidth]{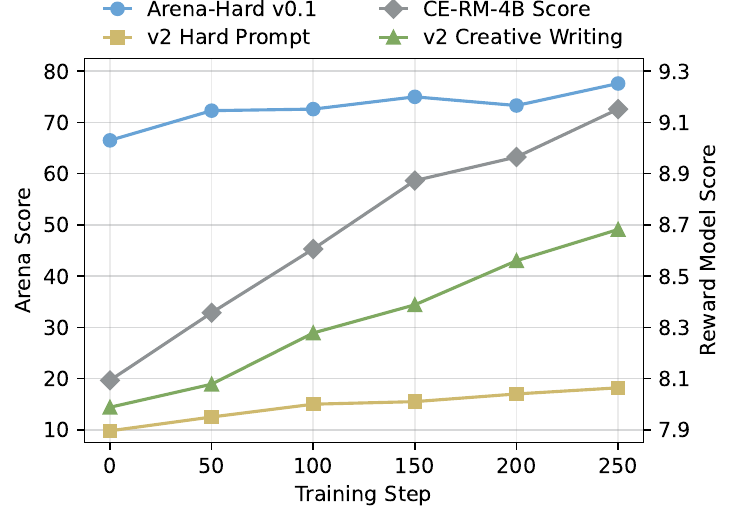}
    \caption{During training, the evolution of performance of the policy model across three evaluation scenarios, along with the averaged evaluation scores that the reward model assigns to the policy rollouts. In the RL practice presented here, the policy model is Qwen3-8B and the reward model is our CE-RM-4B (without test-time scaling), with GRPO and a group size of 8.}
    \label{fig:rlpractice}
\end{figure}

Given that most pointwise GRMs are either not open-sourced or perform poorly, we take as baselines the larger-parameter CompassJudger1-32B and the reward model trained without unified criteria as described in Section~\ref{ablation}. They are compared against our CE-RM-4B, as well as that using test-time scaling at four times. We choose Qwen3-8B as the policy model to be optimized, and all experiments apply the similar training settings as those in Section~\ref{implementation} with a total of 250 training steps. We also examine the impact of different group sizes in GRPO on optimization performance. As shown in Table~\ref{tab:RL_practice} and Figure~\ref{fig:rlpractice}, our model yields stable improvements during RL and significantly outperforms the baseline RMs. Furthermore, both increasing the number of rollout trajectories and applying test-time scaling bring substantial additional gains, enabling the optimized policy model to surpass Qwen3-14B.

\section{Related Work}

The reward model is an important research direction in automatic evaluation, as it provides critical feedback for model alignment. Traditional discriminative reward models are typically trained based on the Bradley–Terry model~\citep{bradley1952rank}, and have been shown to suffer from weak calibration and cross-prompt generalization~\citep{sun2025rethinking, zhanggenerative}. With the rapid development of LLMs, LLM-based generative reward models (GRMs) have emerged, which replace a single scalar score with natural language evaluations. This enables more flexible and informative rewards with better interpretability. In addition, they can leverage test-time scaling to further improve evaluation robustness~\citep{liu2025inference}.

A major challenge in developing GRMs lies in obtaining high-quality supervised evaluation data. Building on existing preference datasets, a common approach is to distill chains of thought from frontier LLMs to enrich the original preference labels~\citep{DBLP:journals/corr/abs-2505-02387, huang2025think, guo2025reward}. Another direction allows the target model itself to generate candidate evaluations, which are then checked with preference labels to perform rejection finetuning in a self-bootstrapping manner~\citep{wang2024self, whitehouse2025j1}. Other studies improve the generalization ability of GRMs by mixing diverse evaluation tasks, including even general instruction-following data, into training~\citep{cao2024compassjudger, wang2025direct}.

In addition, since most preference datasets and generative reward models adopt a pairwise comparison protocol, the correctness of the model's evaluation judgment can be easily verified. Early work often relied on DPO-style optimization, where each preference pair is constructed from a correct and an incorrect evaluation~\citep{mahan2024generative, ye2025learning, yu2025improve}. More recently, driven by the emergence of reinforcement learning with verifiable rewards (RLVR), research has increasingly shifted toward on-policy optimization (e.g., GRPO), further improving the evaluation performance of GRMs~\citep{chen2025judgelrm, DBLP:journals/corr/abs-2505-02835, liu2025inference, whitehouse2025j1}.

\section{Conclusion}

In this work, we start from the limitations of current generative reward models and aim to narrow the gap between their benchmark performance and actual effectiveness in RL practice. To this end, we introduce a two-stage evaluation scheme that enforces unified, query-based criteria across multiple candidate responses. Through a proposed RL method that strengthens the generation of criteria without pointwise label annotations, we train an efficient generative reward model, CE-RM-4B. It features pointwise evaluation and outperforms other much larger pairwise GRMs on common benchmarks and realistic Best-of-N scenarios, while demonstrating better practical RL improvements.

\section*{Limitations}

Although our proposed method optimizes both criteria and evaluation through a two-stage rollout and achieves superior model performance, there remains room for improving the estimation of the reward signal. For instance, we could use a small amount of data annotated with reliable pointwise score labels as anchors, thereby providing precise reward signals for the corresponding evaluations during RL to perform calibration. Moreover, since our evaluation protocol takes the form of a multi-turn conversation, and many current LLMs already possess certain tool-use capabilities, it is natural to train the model to leverage tools for more effective evaluation. In fact, we have made some efforts in this direction and found that tool assistance can indeed yield significant gains in evaluation performance for scenarios such as mathematics, code, and factuality; however, it degrades performance on general chat. Nevertheless, these approaches appear promising and can be further explored in future work.

\section*{Ethical considerations}

This work does not pose any ethical issues. All datasets, open-source LLMs, and API calls used in our work are publicly available. We carefully examine the data used in our work to ensure that it does not contain any personally identifying information or offensive content. And we comply with their respective licenses and use them only for research purposes. In addition, we use an AI assistant to help check the grammatical correctness of this paper.

\bibliography{custom}

\appendix

\section{Evaluation Prompts}
\label{prompts}

We provide our evaluation prompts in Figures~\ref{fig:prompt1}--\ref{fig:prompt3}.

\begin{figure*}[h!]
\centering
\begin{prompt}{The prompt for the first evaluation setting and Eq. (1) in Section 2}
Your task is to evaluate the quality of a response to the given user query. \\

Provide your evaluation with a careful and comprehensive analysis, followed by a corresponding overall quality score from 0 to 10 within $\boxed{\phantom{0}}$. \\

Use integers or half-point increments for the score, with higher numbers representing higher quality. \\

Below are the user query and the response: \\

[Start of Query]

\{instruction\}

[End of Query] \\

[Start of Response]

\{response\}

[End of Response]
\end{prompt}
\vspace{-0.5em}
\caption{The prompt for the first evaluation setting and Eq.~(\ref{eq1}) in Section~\ref{preliminary}.}
\label{fig:prompt1}
\end{figure*}

\begin{figure*}[h!]
\centering
\begin{prompt}{The prompt for the second evaluation setting and Eq. (2) in Section 2}
Your task is to evaluate the quality of a response to the given user query. \\

Begin by carefully analyzing the query to fully understand the user's intent and requirements, and then take into account all common and tangible factors that can indicate the response quality. \\

From these considerations and analyses, derive the final evaluation criteria list between [Start of Criteria] and [End of Criteria], with one criterion per line. \\

Next, for each criterion, focus on its concerns and carefully evaluate the corresponding specific quality of the response, providing the detailed analysis as well as relevant arguments, followed by the corresponding quality score from 0 to 5 within $\boxed{\phantom{0}}$. \\

Finally, based on the analyses of these criteria, including their relative importance and scores, conduct a comprehensive evaluation of the response's overall quality with sufficient and explicit evidence, and then provide a corresponding overall quality score from 0 to 10 within $\boxed{\phantom{0}}$. \\

Use integers or half-point increments for all scores, with higher numbers representing higher quality. \\

Below are the user query and the response: \\

[Start of Query]

\{instruction\}

[End of Query] \\

[Start of Response]

\{response\}

[End of Response]
\end{prompt}
\vspace{-0.5em}
\caption{The prompt for the second evaluation setting and Eq.~(\ref{eq2}) in Section~\ref{preliminary}.}
\label{fig:prompt2}
\end{figure*}

\begin{figure*}[h!]
\centering
\begin{prompt}{The prompts for the third evaluation setting and Eq. (3) in Section 2}
\textbf{Stage 1} \\

Your task is to produce a minimal set of criteria for evaluating the quality of potential responses to the user query given below. \\

Begin by carefully analyzing the query to fully understand the user's intent and requirements, and then take into account all common and tangible factors that can indicate the response quality. \\

From these considerations, derive the final evaluation criteria list, which must adhere to the following requirements: \\

- Each criterion should consist of a concise term as well as its unambiguous description.

- The number of criteria is not necessarily the more the better; Fewer yet comprehensive is more desired.

- The criteria should be sufficient and complete, ensuring that no essential aspects or key signals of response quality are omitted.

- The criteria should be necessary and non-overlapping, with each one indispensable, distinct in perspective, and strictly orthogonal to others. \\

Provide the relevant analysis first, followed by the numbered list of criteria between [Start of Criteria] and [End of Criteria], with one criterion per line and the more important ones coming first. \\

Below is the user query: \\

[Start of Query]

\{instruction\}

[End of Query] \\\\

\textbf{Stage 2} \\

Now that you have a response to the previous user query, your new task is to evaluate it using the criteria list you have produced. \\

For each criterion, focus on its concerns and carefully evaluate the corresponding specific quality of the response, providing the detailed analysis as well as relevant arguments, followed by the corresponding quality score from 0 to 5 within $\boxed{\phantom{0}}$. \\

Moreover, if the response demonstrates strengths or weaknesses beyond the scope of your criteria list, introduce an additional criterion titled "Other Point(s)," discussing them and considering them as bonus points or deductions as appropriate. \\

Finally, based on the analyses of these criteria, including their relative importance and scores, conduct a comprehensive evaluation of the response's overall quality with sufficient and explicit evidence, and then provide a corresponding overall quality score from 0 to 10 within $\boxed{\phantom{0}}$. \\

Use integers or half-point increments for all scores, with higher numbers representing higher quality. \\

Below is the response: \\

[Start of Response]

{response}

[End of Response]
\end{prompt}
\vspace{-0.5em}
\caption{The prompts for the third evaluation setting and Eq.~(\ref{eq3}) in Section~\ref{preliminary}.}
\label{fig:prompt3}
\end{figure*}

\end{document}